\documentclass[letterpaper,10pt,conference]{ieeeconf}
\IEEEoverridecommandlockouts
\usepackage{amsmath,amssymb,amsfonts}
\usepackage{graphicx}
\usepackage{booktabs}
\usepackage{multirow}
\usepackage{array}
\usepackage{cite}
\usepackage{microtype}
\usepackage{float}
\usepackage{placeins}
\usepackage[hidelinks]{hyperref}

\title{SAGE: Safety-Aligned Gradient Enforcement for Human--Robot Collaboration}

\author{Yisen Li$^{*,1}$, Hao Zhang$^{*,1,2}$, Ruize Geng$^{2}$, Yves Tseng$^{1}$, Ding Zhao$^{\dagger,2}$ and H. Eric Tseng$^{\dagger,1}$
\thanks{$^{*}$Equal contribution.}
\thanks{$^{\dagger}$Correspondance to H. Eric Tseng (hongtei.tseng@uta.edu), Ding Zhao (dingzhao@cmu.edu) and Hao E. Zhang (haoz4@andrew.cmu.edu).}
\thanks{$^{1}$Hao Zhang, Yisen Li, Yves Tseng and H. Eric Tseng are with the University of Texas at Arlington.
        (email: haoz4@andrew.cmu.edu; yisen03@upenn.edu; ytseng@duck.com; hongtei.tseng@uta.edu)}%
\thanks{$^{2}$Hao Zhang, Ruize Geng and Ding Zhao are with Carnegie Mellon University
        (haoz4@andrew.cmu.edu; rgeng3@jh.edu; dingzhao@cmu.edu)}%
}

\begin{document}
\maketitle
\thispagestyle{empty}
\pagestyle{empty}

\begin{abstract}
Multi-party human--robot collaboration poses a dual challenge: robot decisions should remain interpretable and auditable, while executed actions must satisfy safety constraints during physical interaction. Combining explainable decision-tree policies with control-barrier-function (CBF) filtering provides a promising architecture but creates two learning mismatches in multi-agent reinforcement learning. Safety projection changes the action applied to the environment, while the coupled proposal graph can misalign independently optimized actor updates with a team-level update. We present safety-aligned gradient enforcement (SAGE) to address both mismatches. Its shield-annealed internalization layer (SAIL) uses a differentiable finite-penalty proposal map while retaining the exact CBF quadratic program for execution, preserving constraint-normal sensitivity to internalize repeatedly active safety constraints. Team-averaged Lyapunov policy optimization (TALO) constructs a team-aware update reference and applies a Lyapunov half-space correction to regulate independent actor updates. Physical experiments with two humanoid robots and a human partner demonstrate deployment feasibility. Across nine simulation scenarios, SAGE achieves a 71.0\% success rate with 0.5 collision steps per thousand environment steps. Ablations show that direct CBF filtering reduces collision frequency by 98.5\% but decreases success from 67.3\% to 59.3\%. SAIL reduces proposal violation by 48.8\% and proposal--execution correction by 85.2\%, while TALO reduces the update-consistency gap by 50.8\%.
\end{abstract}

\section{Introduction}
\label{sec:introduction}

Physical human--robot collaboration increasingly involves multiple participants acting through a shared object rather than performing loosely coupled individual tasks \cite{zhang2026interaction}. Cooperative transport is representative of this regime: each local action changes the payload motion and, through rigid-body coupling, the feasible responses and safety margins of the other carriers \cite{bernardtiong2024cooperative}. Human participation adds another challenge, since robot decisions should remain inspectable during interaction rather than inferred only from the final task outcome \cite{ajoudani2018progress}. A practical controller must therefore combine decentralized decision-making, joint safety enforcement, and decision-level auditability \cite{dragan2013legibility}.

Multi-agent reinforcement learning (MARL) supports decentralized policies under shared objectives, while centralized-training decentralized-execution methods support heterogeneous agents and long-horizon coordination \cite{yu2022surprising}. Decision trees provide explicit observation-to-action maps \cite{bastani2018viper}, whereas control barrier functions (CBFs) enforce state-dependent constraints through online quadratic programs \cite{ames2017control}. Their combination separates the auditable nominal decision from the constraint-enforced executed action. This separation creates a learning mismatch: repeated safety intervention causes the action generating the return to differ from the nominal proposal, allowing the policy to remain dependent on downstream correction rather than internalizing repeatedly active constraints\cite{cheng2019end}.

\begin{figure}[t]
    \centering
    \includegraphics[width=\columnwidth]{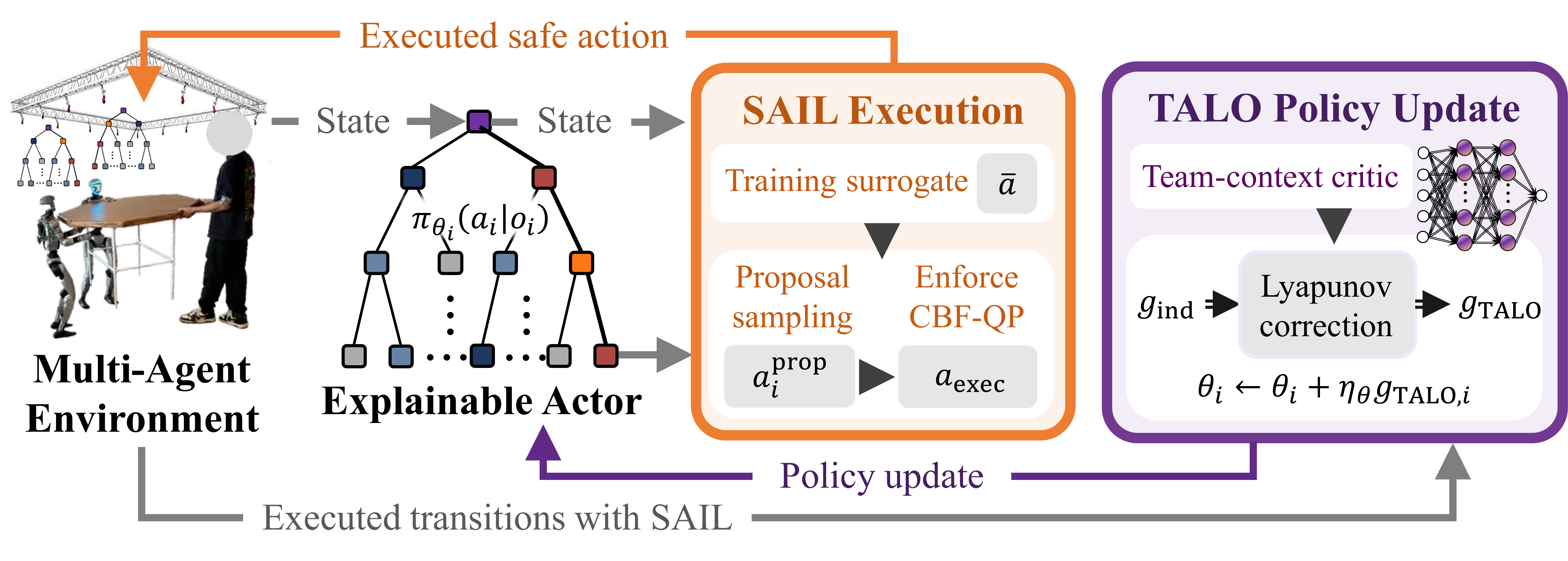}
    \caption{Architecture of SAGE integrating auditable hard-tree nominal policies, SAIL-based proposal learning, exact joint CBF-QP execution, and TALO-based multi-actor policy optimization.}
    \label{fig:sage_intro}
\end{figure}

This mismatch also appears in the policy gradient. Near an active linear constraint, the Jacobian of an exact Euclidean projection removes sensitivity along the constraint-normal direction, weakening the proposal-side signal that would otherwise reduce repeated intervention \cite{kuba2022trust}. In shared-payload coordination, this effect is further coupled because one carrier's proposal can alter the active safety geometry and the correction applied to multiple carriers. A second mismatch arises in parameter space \cite{milani2022maviper}. Independently parameterized actors are typically optimized through per-agent clipped proximal-policy objectives, whereas the shared reward and joint proposal distribution also induce a team-level update \cite{zhong2024harl}. These update fields need not coincide, particularly when the proposal computation itself couples agents through a joint safety-aware transformation \cite{zhang2026halo}. The resulting problem is to learn around the execution filter while maintaining consistency between local actor updates and the team behavior induced by the coupled training graph.

Three-carrier transport provides a compact setting in which these couplings extend beyond a single interacting pair \cite{paleja2022icct}. A local action can simultaneously affect multiple partners and safety margins, while different internal-force patterns can remain compatible with the same payload wrench \cite{Cognition2Control}. The resulting closed interaction cycle captures multi-partner physical and learning dependencies without requiring a large-team formulation, making it a suitable setting for studying decentralized policy learning under coupled safety constraints and team-level coordination \cite{xiao2023barriernet}.

We develop safety-aligned gradient enforcement (SAGE) for this architecture, as summarized in Fig.~\ref{fig:sage_intro}. Each robot uses an independently parameterized hard oblique decision tree to generate a local nominal action, and a joint CBF-QP enforces the executed constraints. The shield-annealed internalization layer (SAIL) introduces a finite-penalty proposal map only in the training graph so that constraint-normal sensitivity is retained. Team-averaged Lyapunov policy optimization (TALO) addresses the complementary multi-actor update problem by constructing a team-aware reference from a team-averaged representation and the joint proposal likelihood. The contributions are summarized as follows: 1) SAGE formulates a safety-constrained MARL architecture with auditable hard-tree nominal policies explicitly separated from exact joint CBF execution; 2) SAIL provides a differentiable internalization mechanism for the tree--CBF interface, while TALO introduces a team-aware update reference that regulates independently optimized actors without changing their deployed information structure; and 3) comprehensive experiments demonstrate improved safety--performance tradeoffs, reduced nominal dependence on CBF correction, better team-update consistency, and auditable multi-agent decisions.

\begin{figure*}[t]
    \centering
    \includegraphics[width=\textwidth]{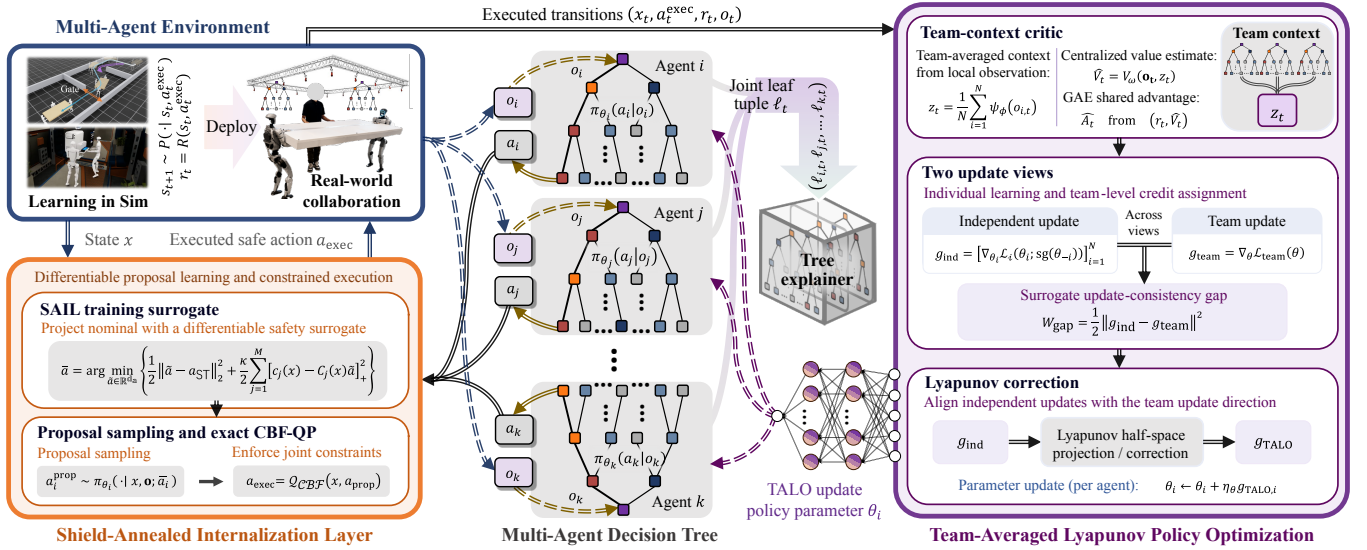}
    \caption{Training and deployment structure of SAGE, where hard-tree actors generate auditable nominal actions, SAIL forms differentiable proposal centers for safety-aware proposal learning, the exact joint CBF-QP maps sampled proposals to constraint-satisfying executed actions, and TALO uses executed-transition returns and a team-aware update reference to regulate independent actor updates, while SAIL, exploration, the critic, and TALO are removed at deployment, leaving only the learned hard-tree actors and the exact joint CBF-QP.}
    \label{fig:sage_framework}
\end{figure*}

\section{Related Work}
\label{sec:related_work}

\subsection{Multi-Agent Transport and Human--Robot Collaboration}

Cooperative transport has been studied through model-based coordination, interaction-force regulation, and learning-based control, particularly when contact distribution, partner response, and long-horizon coordination are difficult to model \cite{zhang2025multi}. Bernard-Tiong et al. use ternary interaction-force representations for two-robot transport in simulation and hardware \cite{bernardtiong2024cooperative}, while prior human--robot collaboration work studies shared-payload human--humanoid transport with independently parameterized policies \cite{Cognition2Control}. MAPPO provides a strong cooperative baseline, while HAPPO and HARL support heterogeneous policies and sequential updates \cite{yu2022surprising,kuba2022trust,zhong2024harl}. These methods enable coordination but do not address the training mismatch introduced when auditable nominal policies are followed by a safety filter.

\subsection{Auditable Policies for Robot Decision-Making}

Decision trees expose routing conditions and leaf actions, enabling direct inspection of deployed nominal policies. VIPER distills neural policies into decision trees \cite{bastani2018viper}, while MAVIPER extends tree extraction to multi-agent settings \cite{milani2022maviper}. Directly optimized alternatives include interpretable continuous control trees with differentiable routing \cite{paleja2022icct}, MIXRTs with recurrent soft trees and value mixing \cite{liu2025mixrts}, and DTPO with direct policy-gradient optimization of discrete trees \cite{vos2024dtpo}. Our focus is decision-level auditability: the nominal policy exposes its active predicates and local action, while safety is enforced separately at execution.

\subsection{Safe Learning and Coupled Multi-Agent Optimization}

CBFs convert state-safety requirements into QP-enforced action constraints \cite{ames2017control}. They have been combined with model-free reinforcement learning \cite{cheng2019end}; OptNet and BarrierNet integrate differentiable constrained optimization with policy learning \cite{zhang2025bi,xiao2023barriernet}, while graph-based methods extend barrier learning to larger multi-agent systems \cite{zhang2025gcbfplus}. Exact projection, however, can suppress proposal-side sensitivity normal to active constraints; SAIL retains the exact QP for execution and uses a finite-penalty map for proposal learning. Coupled policy updates pose a separate problem. Differentiable-game analysis characterizes interacting gradient fields, LOLA differentiates through anticipated opponent updates \cite{balduzzi2018mechanics,foerster2018lola}, and heterogeneous-agent Lyapunov optimization has been applied to HRC \cite{zhang2026halo}. In our setting, independently clipped actor updates can differ from the team update induced by the joint proposal likelihood, especially under a coupled proposal map. TALO addresses this mismatch with a team-aware reference and local Lyapunov half-space correction.

\section{Methodology}
\label{sec:method}

\subsection{Problem Formulation}
\label{sec:problem}

We model cooperative transport as a decentralized partially observable Markov decision process
\begin{equation}
\mathcal M=\left\langle
\mathcal S,\{\mathcal O_i,\mathcal A_i\}_{i=1}^N,
P,R,\gamma
\right\rangle ,
\label{eq:decpomdp}
\end{equation}
where $s_t\in\mathcal S$ is the system state, carrier $i$ observes $o_{i,t}\in\mathcal O_i$, $\mathcal A_i\subset\mathbb R^{d_{a,i}}$ is its bounded high-level action space, $P$ is the transition kernel, $R$ is the shared reward, and $\gamma\in(0,1)$ is the discount factor. Each carrier has an independently parameterized nominal policy $\mu_{\theta_i}:\mathcal O_i\rightarrow\mathcal A_i$ with no parameter sharing. We write $\theta=[\theta_1^\top,\ldots,\theta_N^\top]^\top$, $\mathcal A=\prod_i\mathcal A_i$, and $d_a=\sum_i d_{a,i}$. During training, the nominal actors induce a stochastic joint proposal $a_t^{\mathrm{prop}}\in\mathcal A$, while the safety layer produces the executed action $a_t^{\mathrm{exec}}\in\mathcal A$. The shared objective is evaluated on executed transitions,
\begin{equation}
J(\theta)=\mathbb E\!\left[
\sum_{t=0}^\infty \gamma^t R(s_t,a_t^{\mathrm{exec}})
\right].
\label{eq:team_return}
\end{equation}
Let $x_t=\Phi(s_t)$ collect the physical variables required by the safety model. The execution layer enforces a set of jointly coupled affine safety rows
\begin{equation}
C(x_t)a_t^{\mathrm{exec}}\ge c(x_t),
\label{eq:joint_safety_rows}
\end{equation}
where $C(x_t)\in\mathbb R^{M\times d_a}$, $c(x_t)\in\mathbb R^M$, and $M$ is the number of enforced safety rows obtained from the one-step CBF conditions. The learning problem is to maximize $J(\theta)$ with decentralized nominal policies while the action actually applied to the environment is generated by this joint constrained execution map. This separation creates the proposal--execution mismatch addressed by SAIL and the independent--team update mismatch addressed by TALO. Fig.~\ref{fig:sage_framework} summarizes the complete training and deployment.

\subsection{Hard-Tree Policies with Exact Joint CBF Execution}
\label{sec:tree_cbf}

Carrier $i$ represents its nominal policy with a hard oblique decision tree. Omitting the carrier index for readability, internal node $k$ uses a normalized split vector $w_k$ and threshold $b_k$,
\begin{equation}
g_k(o)=\mathbb I[w_k^\top o>b_k],\qquad \|w_k\|_2=1.
\label{eq:hard_gate}
\end{equation}
The hard route selects a leaf $\ell_\star(o)$. Leaf $\ell$ stores an affine controller $K_\ell o+\beta_\ell$, and the bounded nominal action is
\begin{equation}
a_{\mathrm{nom}}(o)
=c_a+s_a\odot\tanh\!\left(K_{\ell_\star(o)}o+\beta_{\ell_\star(o)}\right),
\label{eq:hard_tree_policy}
\end{equation}
where $c_a$ and $s_a$ are the center and half-range of the action bounds and $\odot$ denotes elementwise multiplication. The deployed decision is therefore a deterministic root-to-leaf predicate sequence followed by an explicit affine leaf command. Hard routing is retained in the forward pass. To optimize split parameters, we use a straight-through estimator with soft backward gate
\begin{equation}
\begin{aligned}
g_k^{\mathrm{ST}}&=\operatorname{sg}(g_k-p_k)+p_k,\\
p_k&=\left[1+\exp\!\left(-\frac{w_k^\top o-b_k}{\tau}\right)\right]^{-1}.
\end{aligned}
\label{eq:st_gate}
\end{equation}
where $\operatorname{sg}(\cdot)$ denotes stop gradient and $\tau>0$ is the backward temperature, which is scheduled during training while the forward route remains hard.

For execution, the stacked proposal is projected onto the joint constraint-admissible action set by
\begin{equation}
\mathcal Q_{\mathrm{CBF}}(x,a)
=\arg\min_{\tilde a\in\mathcal A}
\frac12\|\tilde a-a\|_2^2
\quad \text{s.t.}\quad
C(x)\tilde a\ge c(x).
\label{eq:qpshield}
\end{equation}
When the feasible set is nonempty, the strictly convex objective gives a unique filtered action satisfying the enforced affine rows. All constraint-satisfaction statements below are conditional on this feasibility condition. The QP is solved directly for every rollout and deployment step; it is not replaced by a learned safety model. Here, ``exact'' refers to solving this constrained projection directly rather than using the SAIL surrogate; it does not imply an exact dynamics predictor. The CBF statement is predictor-level: for the one-step predictor $\hat x^+=\hat F(x)+\hat B(x)a$, each affine row is assumed to be a conservative lower bound on the corresponding discrete-time barrier condition. A bounded one-step model error can be incorporated into the row margin; without such a bound, no model-free forward-invariance claim is made. If one carrier slot is occupied by a human partner, its measured high-level action is treated as a fixed exogenous block and the QP optimizes only the robot action blocks.

\subsection{Shield-Annealed Internalization Layer}
\label{sec:sail}

Let $a_{\mathrm{ST}}$ denote the stacked hard-forward nominal action. Differentiating directly through the exact projection in Eq.~\eqref{eq:qpshield} is undesirable near active constraints because the projection removes normal sensitivity. SAIL instead forms only the training proposal center through
\begin{equation}
\begin{aligned}
\bar a&=\Pi_\kappa(x,a_{\mathrm{ST}})\\
&=\arg\min_{\tilde a\in\mathbb R^{d_a}}
\Bigg\{
\frac12\|\tilde a-a_{\mathrm{ST}}\|_2^2\\
&\hspace{2.5em}
+\frac{\kappa}2\sum_{j=1}^M
[c_j(x)-C_j(x)\tilde a]_+^2
\Bigg\}.
\end{aligned}
\label{eq:sail_projection}
\end{equation}
where $[z]_+=\max(z,0)$ and $\kappa>0$ is the penalty coefficient. The first term makes the objective strongly convex, so the ideal minimizer is unique. In implementation, a fixed number of damped Newton steps is unrolled and differentiated; away from active-set switching surfaces, the squared-hinge objective is piecewise quadratic and the Newton system uses the current positive-residual rows. Each nonzero safety row is normalized before entering SAIL so that the penalty magnitude does not depend on row units. The penalty coefficient $\kappa$ is scheduled during training to progressively reduce the soft-constraint residual. The Jacobian statements below characterize the ideal minimizer; the implemented finite-step SAIL map is differentiated directly and is not assumed to satisfy them exactly. The distinction from direct QP differentiation can be seen locally. For a fixed linearly independent row set $\mathcal J$ that is active for the exact projection and has positive penalty residual in SAIL, with no active action-box face, the exact projection has Jacobian
\begin{equation}
\frac{\partial\mathcal Q_{\mathrm{CBF}}}{\partial a}
=I-C_\mathcal J^\top(C_\mathcal J C_\mathcal J^\top)^{-1}C_\mathcal J ,
\label{eq:qp_jacobian}
\end{equation}
which annihilates directions in the active-row space. In the corresponding SAIL region,
\begin{equation}
\frac{\partial\Pi_\kappa}{\partial a_{\mathrm{ST}}}
=(I+\kappa C_\mathcal J^\top C_\mathcal J)^{-1},
\label{eq:sail_jacobian}
\end{equation}
which remains positive definite for finite $\kappa$. SAIL therefore attenuates, rather than removes, constraint-normal sensitivity. For one normalized active row, the normal component is scaled by $1/(1+\kappa)$.

Each carrier samples a bounded proposal $a_i^{\mathrm{prop}}\sim\pi_{\theta_i}(\cdot\mid x,\mathbf o;\bar a_i)$ with fixed support contained in $\mathcal A_i$. Conditioned on $(x,\mathbf o,\bar a)$, the proposal noises are sampled independently across carriers, so the joint proposal density factorizes; positive density in the support interior makes the likelihood ratio at stored proposals well-defined. The training chain is
\begin{equation}
a_{\mathrm{ST}}
\xrightarrow{\Pi_\kappa}\bar a
\xrightarrow{\pi_\theta}a_{\mathrm{prop}}
\xrightarrow{\mathcal Q_{\mathrm{CBF}}}a_{\mathrm{exec}}.
\label{eq:action_chain}
\end{equation}
The policy likelihood is evaluated at $a_{\mathrm{prop}}$, whereas rewards, next states, and critic targets are generated from $a_{\mathrm{exec}}$. The exact CBF-QP remains the sole mechanism enforcing the executed affine constraints.

\subsection{Team-Averaged Lyapunov Policy Optimization}
\label{sec:talo}

SAIL addresses the action-space mismatch but does not determine how independently optimized actors should relate to the team objective. TALO introduces a team-aware update reference. A shared encoder $\psi_\phi$ maps each local observation to a latent vector, and the team-averaged context
\begin{equation}
z_t=\frac1N\sum_{i=1}^N\psi_\phi(o_{i,t}),\qquad
\widehat V_t=V_\omega(\mathbf o_t,z_t),
\label{eq:team_context}
\end{equation}
provides a permutation-insensitive summary to the centralized critic. Here $\mathbf o_t=[o_{1,t}^\top,\ldots,o_{N,t}^\top]^\top$, while $\phi$ and $\omega$ are encoder and critic parameters. The shared advantage $\widehat A_t$ is computed from executed transitions using generalized advantage estimation.

Let $\rho_i(\theta)$ be the proposal likelihood ratio of carrier $i$ with respect to the behavior policy. Since the joint proposal density factorizes, its likelihood ratio is $\rho_{\mathrm{team}}=\prod_i\rho_i$. With clip range $\epsilon_{\mathrm{clip}}>0$, define the clipped proximal-policy objective $\ell(\rho,A)=\min(\rho A,\operatorname{clip}(\rho,1-\epsilon_{\mathrm{clip}},1+\epsilon_{\mathrm{clip}})A)$. The independent and team-reference surrogates are
\begin{equation}
L_i=\mathbb E_t[\ell(\rho_i,\widehat A_t)],
\qquad
L_{\mathrm{team}}=\mathbb E_t[\ell(\rho_{\mathrm{team}},\widehat A_t)].
\label{eq:ppo_surrogates}
\end{equation}
When $\nabla_{\theta_i}L_i$ is formed, partner nominal branches are detached before the SAIL map; $L_{\mathrm{team}}$ retains the full composite graph. We then define
\begin{equation}
\begin{aligned}
g_{\mathrm{ind}}
&=\begin{bmatrix}
\nabla_{\theta_1}L_1\\ \vdots\\ \nabla_{\theta_N}L_N
\end{bmatrix},
\qquad
g_{\mathrm{team}}=\nabla_\theta L_{\mathrm{team}},\\
W_{\mathrm{gap}}
&=\frac12\|g_{\mathrm{ind}}-g_{\mathrm{team}}\|_2^2 .
\end{aligned}
\label{eq:talo_gap}
\end{equation}

Here, $g_{\mathrm{team}}$ is a local team-aware reference, not an assumed globally optimal policy gradient; $W_{\mathrm{gap}}$ therefore measures update inconsistency rather than distance to an optimal direction. The Lyapunov construction then acts as a rectification operator: the closest update to $g_{\mathrm{ind}}$ satisfying the local descent condition
\begin{equation}
\nabla_\theta W_{\mathrm{gap}}^\top g
\le -\sigma W_{\mathrm{gap}},
\qquad \sigma>0,
\label{eq:talo_condition}
\end{equation}
is implemented with the regularized correction
\begin{equation}
g_{\mathrm{TALO}}
=g_{\mathrm{ind}}
-\frac{[\Delta_{\mathrm{gap}}]_+}
{\|\nabla_\theta W_{\mathrm{gap}}\|_2^2+\epsilon_{\mathrm{TALO}}}
\nabla_\theta W_{\mathrm{gap}},
\label{eq:talo_projection}
\end{equation}
where $\Delta_{\mathrm{gap}}=\nabla_\theta W_{\mathrm{gap}}^\top g_{\mathrm{ind}}+\sigma W_{\mathrm{gap}}$ and $\epsilon_{\mathrm{TALO}}>0$ is a numerical regularizer. The actor parameters are updated by $\theta\leftarrow\theta+\eta_\theta g_{\mathrm{TALO}}$, where $\eta_\theta>0$ is the actor step size. Without denominator regularization, the projected field satisfies the local descent condition in Eq.~\eqref{eq:talo_condition} at differentiable minibatch points whenever $\nabla_\theta W_{\mathrm{gap}}\neq0$. This is a local property of the frozen-minibatch surrogate and does not imply global convergence of stochastic proximal-policy optimization. Hessian--vector products needed for $\nabla_\theta W_{\mathrm{gap}}$ are obtained by automatic differentiation without forming an explicit Hessian. SAIL, the critic, exploration, and TALO are removed at deployment; only the learned hard-tree actors and the exact joint CBF-QP remain.

\section{Results and Discussion}
\label{sec:experiments}

\subsection{Experimental Setup}
\label{sec:experimental_setup}

\begin{figure}[t]
    \centering
    \includegraphics[width=\columnwidth]{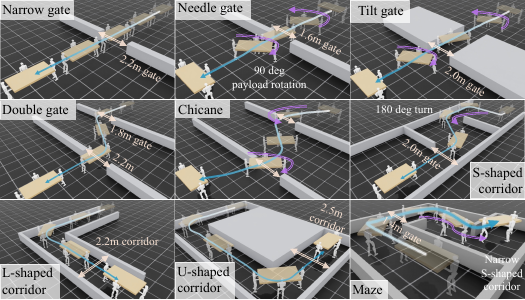}
    \caption{Nine multi-carrier cooperative-transport scenarios grouped by row into LCP (Narrow Gate, Needle Gate, Tilt Gate), STR (Double Gate, Chicane, S-Shaped Corridor), and CCN (L-Corridor, U-Shaped Corridor, Maze), with overlaid payload poses indicating representative trajectory configurations.}
    \label{fig:nine_scenarios}
\end{figure}

We evaluate nine three-carrier simulation scenarios in Isaac Lab~\cite{mittal2023orbit} (Fig.~\ref{fig:nine_scenarios}). They cover local clearance and passage (LCP), sequential turning and reorientation (STR), and constrained corridor navigation (CCN). Three non-collinear grasps couple the carriers through a shared payload. Each carrier outputs $a_i=[\Delta x_i,\Delta y_i,\Delta\psi_i]^\top\in[-1,1]^3$ through a frozen whole-body controller. HAPPO is the unshielded reference. HAPPO+CBF adds exact joint filtering, TALO+CBF and SAIL+CBF add the respective training mechanisms, and SAGE+CBF combines both. In TALO+CBF, the SAIL proposal-center map is replaced by the identity map; in SAIL+CBF, TALO is omitted. All shielded variants use the same exact CBF-QP for executed actions; SAIL, TALO, the critic, and exploration are training-only. Unless otherwise stated, reported $\pm$ values and error bars denote mean $\pm$ standard deviation across evaluation runs.

\begin{table*}[!t]
\caption{Performance and safety across the LCP, STR, and CCN cooperative-transport task families, with SR reported in \% and Col./1k in collision steps per thousand environment steps.}
\label{tab:global_performance}
\centering

\small
\renewcommand{\arraystretch}{1.10}

\begin{tabular*}{\textwidth}
{@{\extracolsep{\fill}}ll*{5}{cc}@{}}
\toprule

\multirow{2}{*}{Category}
& \multirow{2}{*}{Scenario}
& \multicolumn{2}{c}{HAPPO}
& \multicolumn{2}{c}{HAPPO+CBF}
& \multicolumn{2}{c}{TALO+CBF}
& \multicolumn{2}{c}{SAIL+CBF}
& \multicolumn{2}{c}{SAGE+CBF} \\

\cmidrule(lr){3-4}
\cmidrule(lr){5-6}
\cmidrule(lr){7-8}
\cmidrule(lr){9-10}
\cmidrule(lr){11-12}

&
& SR$\uparrow$
& Col./1k$\downarrow$
& SR$\uparrow$
& Col./1k$\downarrow$
& SR$\uparrow$
& Col./1k$\downarrow$
& SR$\uparrow$
& Col./1k$\downarrow$
& \multicolumn{2}{c}
  {SR$\uparrow$\hspace{1.7em}Col./1k$\downarrow$} \\

\midrule

\multirow{4}{*}{LCP}
& Narrow Gate
& 76.7 & 156.7
& 73.0 & 0.2
& 74.0 & 0.4
& 74.0 & 0.4
& 76.0 & 0.1 \\

& Needle Gate
& 68.7 & 42.2
& 48.1 & 0.6
& 48.3 & 1.8
& 54.6 & 0.1
& 58.3 & 0.1 \\

& Tilt Gate
& 68.3 & 235.4
& 55.7 & 1.1
& 78.7 & 0.1
& 74.8 & 1.3
& 77.7 & 0.6 \\

\cmidrule(lr){2-12}

& LCP Mean
& 71.2 & 144.8
& 58.9 & 0.6
& 67.0 & 0.8
& 67.8 & 0.6
& 70.7 & 0.3 \\

\midrule

\multirow{4}{*}{STR}
& Double Gate
& 66.3 & 147.7
& 64.3 & 0.3
& 67.0 & 0.6
& 70.6 & 0.2
& 69.7 & 0.3 \\

& Chicane
& 63.3 & 85.1
& 59.3 & 1.5
& 74.3 & 0.8
& 64.8 & 0.8
& 78.0 & 1.0 \\

& S-Shaped Corridor
& 80.0 & 52.4
& 67.7 & 1.1
& 71.0 & 0.1
& 79.0 & 0.6
& 75.3 & 0.5 \\

\cmidrule(lr){2-12}

& STR Mean
& 69.9 & 95.1
& 63.8 & 1.0
& 70.8 & 0.5
& 71.5 & 0.5
& 74.3 & 0.6 \\

\midrule

\multirow{4}{*}{CCN}
& L-Corridor
& 70.3 & 85.2
& 71.3 & 10.3
& 60.0 & 4.4
& 72.8 & 5.2
& 75.0 & 1.0 \\

& U-Shaped Corridor
& 48.7 & 72.7
& 46.7 & 0.4
& 74.3 & 3.1
& 59.4 & 3.4
& 62.3 & 0.2 \\

& Maze
& 63.2 & 270.3
& 47.3 & 1.9
& 54.6 & 1.2
& 59.7 & 0.5
& 66.3 & 0.3 \\

\cmidrule(lr){2-12}

& CCN Mean
& 60.7 & 142.7
& 55.1 & 4.2
& 63.0 & 2.9
& 64.0 & 3.0
& 67.9 & 0.5 \\

\bottomrule
\end{tabular*}

\vspace{0.9em}

\small
\renewcommand{\arraystretch}{1.15}

\begin{tabular*}{\textwidth}
{@{\extracolsep{\fill}}lcccccc@{}}
\toprule

\multicolumn{6}{c}{
Overall Performance and Safety Analysis
} \\

\midrule

Method
& SR (\%) $\uparrow$
& Return $\uparrow$
& Col./1k $\downarrow$
& DR (\%) $\downarrow$
& Proposal viol. (\%) $\downarrow$ \\

\midrule

HAPPO
& $67.3\pm4.3$
& $4.23\pm0.08$
& $127.5\pm65.6$
& $16.0\pm1.8$
& N/A \\

HAPPO+CBF
& $59.3\pm6.4$
& $3.69\pm0.12$
& $1.9\pm3.3$
& $37.4\pm4.5$
& $24.2\pm14.7$ \\

TALO+CBF
& $66.9\pm4.6$
& $4.05\pm0.11$
& $1.4\pm1.5$
& $27.7\pm4.6$
& $19.3\pm8.9$ \\

SAIL+CBF
& $67.7\pm4.3$
& $4.13\pm0.10$
& $1.4\pm1.9$
& $25.6\pm3.2$
& $12.4\pm5.7$ \\

SAGE+CBF
& $71.0\pm2.7$
& $4.25\pm0.09$
& $0.5\pm0.4$
& $20.8\pm2.3$
& $14.2\pm7.2$ \\

\bottomrule
\end{tabular*}

\end{table*}

\begin{figure}[t]
    \centering
    \includegraphics[width=0.99\columnwidth]{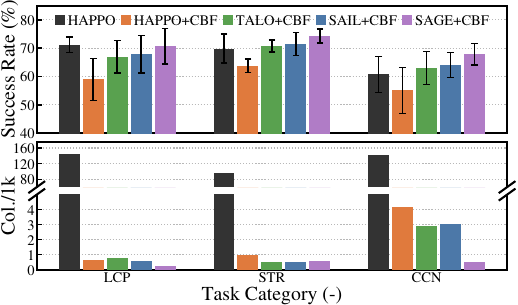}
    \caption{Suite-level performance across LCP, STR, and CCN, with success rate (SR, $\uparrow$) above and collision steps per thousand environment steps (Col./1k, $\downarrow$) below; the broken lower axis accommodates the much larger collision rate of unshielded HAPPO, and error bars denote standard deviation across evaluation runs.}
    \label{fig:main_sr}
\end{figure}

\begin{figure}[t]
    \centering
    \includegraphics[width=0.99\columnwidth]{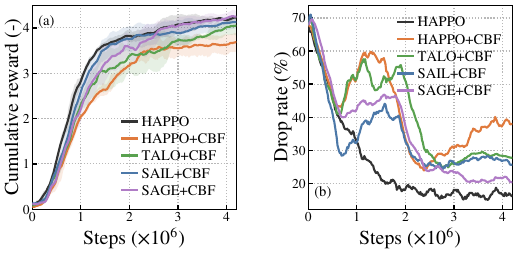}
    \caption{Training dynamics of the five methods showing (a) cumulative reward and (b) payload drop rate, with solid curves denoting the mean and shaded regions the variability across runs.}
    \label{fig:reward_drop}
\end{figure}

\begin{table}[t]
\caption{Association of individual and joint leaf states with route progress in the Isaac Lab Maze.}
\label{tab:leaf_interpretability}
\centering
\small
\renewcommand{\arraystretch}{1.15}
\setlength{\tabcolsep}{8pt}

\begin{tabular}{lccc}
\toprule
Representation
& Purity $\uparrow$
& Shuffle $z$ $\uparrow$
& NMI $\uparrow$ \\
\midrule

Individual leaf
& 0.580
& 97.9
& 0.52 \\

Joint tuple $(\ell_1,\ell_2,\ell_3)$
& 0.845
& 95.2
& 0.78 \\

\bottomrule
\end{tabular}
\end{table}

\begin{figure}[!t]
    \centering
    \includegraphics[width=\linewidth]{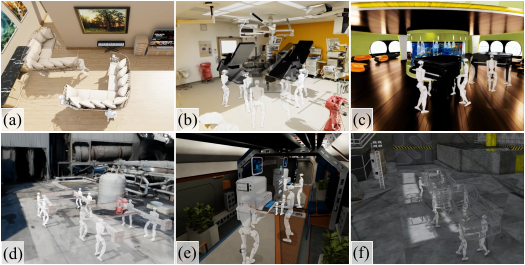}
    \caption{Illustrative multi-robot cooperative-transport scenarios across indoor environments and payload geometries.}
    \label{fig:simulation}
\end{figure}

Task performance is measured by success rate (SR), return, collision steps per thousand environment steps (Col./1k), and payload drop rate (DR). For shielded variants, proposal violation rate is the fraction of raw hard-tree nominal actions $a_{\mathrm{ST}}$ violating at least one affine safety row; the unshielded HAPPO reference is reported as N/A for this constraint-probe metric. Optimization behavior is evaluated through $W_{\mathrm{gap}}$ and the cosine alignment between $g_{\mathrm{ind}}$ and $g_{\mathrm{team}}$. Policy auditability is evaluated from individual- and joint-leaf occupancy, route-progress purity, normalized mutual information (NMI), and shuffle-based $z$ scores.

\subsection{Overall Performance and Safety--Performance Tradeoff}
\label{sec:overall_performance}

Table~\ref{tab:global_performance} and Fig.~\ref{fig:main_sr} show that adding joint CBF filtering to HAPPO reduces collision frequency from $127.5$ to $1.9$ steps per thousand ($98.5\%$), but lowers SR from $67.3\%$ to $59.3\%$. TALO+CBF and SAIL+CBF recover SR to $66.9\%$ and $67.7\%$, respectively. SAGE+CBF reaches $71.0\%$ SR with $0.5$ collision steps per thousand. Relative to HAPPO+CBF, this recovers $11.7$ percentage points of SR and reduces DR from $37.4\%$ to $20.8\%$. These results show that the proposed training mechanisms recover task performance without relaxing the common execution-time safety filter. The nonzero collision counts are consistent with the predictor-level scope in Sec.~\ref{sec:tree_cbf}, rather than a claim of model-free forward invariance.

Fig.~\ref{fig:reward_drop} shows the same trend: direct shielding degrades cumulative reward and increases payload drops, while SAIL and TALO progressively recover performance. SAGE retains this recovery without sacrificing the low-collision regime established by the CBF-QP.

Fig.~\ref{fig:simulation} illustrates cooperative handling of varied furniture and equipment under different workspace constraints.

\begin{figure}[t]
    \centering
    \includegraphics[width=\columnwidth]{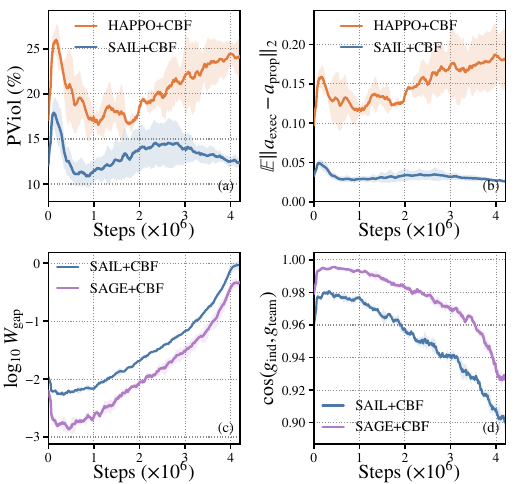}
    \caption{SAIL and TALO mechanism dynamics showing (a) proposal violation rate of the raw hard-tree nominal action $a_{\mathrm{ST}}$, (b) proposal--execution correction, (c) update-consistency gap $W_{\mathrm{gap}}$, and (d) cosine alignment between $g_{\mathrm{ind}}$ and $g_{\mathrm{team}}$.}
    \label{fig:sail_talo_ablation}
\end{figure}

\begin{figure}[t]
    \centering
    \includegraphics[width=0.97\columnwidth]{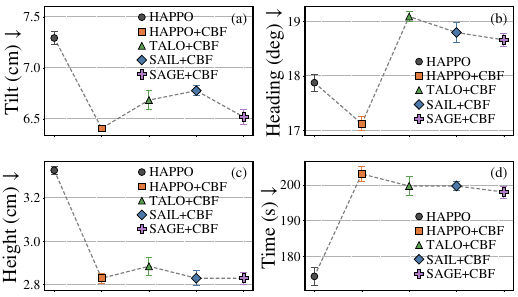}
    \caption{Execution-level comparison across the five methods in (a) payload-tilt-induced vertical deviation, (b) mean step-wise heading error $|\mathrm{yaw}-\mathrm{yaw}_{\mathrm{goal}}|$, (c) payload height error, and (d) task completion time, with error bars indicating variability across runs.}
    \label{fig:payload_panels}
\end{figure}

\begin{figure*}[t]
    \centering
    \includegraphics[width=\textwidth]{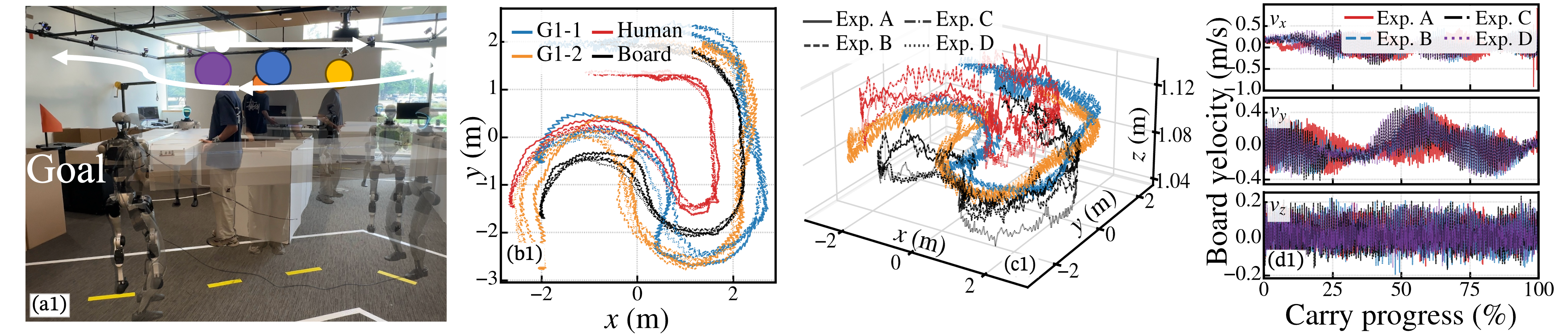}
    \includegraphics[width=\textwidth]{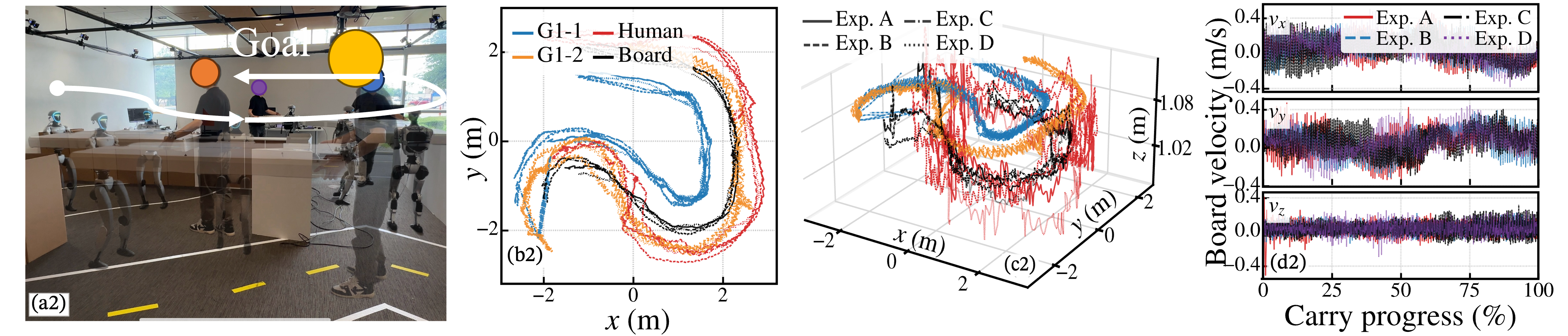}
    \caption{Human--humanoid transport with two Unitree G1s and a human partner in outbound (human on the long edge, G1s at the short ends) and return (human at one short end, G1s at the opposite short end and long edge) formations, showing (a) multi-exposure trial views, (b) top-view and (c) 3-D Vicon trajectories over four trials (Exp.~A--D), and (d) board-corner velocity components versus normalized carry progress.}
    \label{fig:real_world_three_cases}
\end{figure*}

\begin{figure}[t]
    \centering
    \includegraphics[width=\columnwidth]{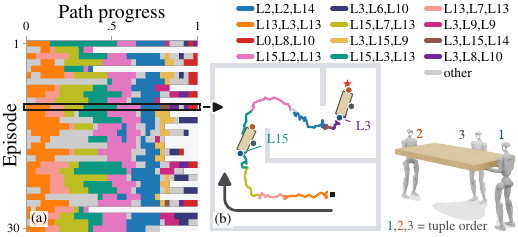}
    \caption{Joint-leaf-tuple audit in the Isaac Lab Maze showing (a) occupancy over path progress across 30 evaluation episodes and (b) the highlighted episode colored by the 12 most frequent tuples, with remaining tuples grouped as other; the black square and red star mark the start and goal, and $1$--$3$ denote tuple order.}
    \label{fig:colortrace}
\end{figure}

\begin{table}[t]
\caption{Transport quality metrics for the real-world deployment.}
\label{tab:realworld_formations}
\centering
\footnotesize
\renewcommand{\arraystretch}{1.15}
\setlength{\tabcolsep}{2.4pt}

\begin{tabular*}{\columnwidth}
{@{\extracolsep{\fill}}lccccc@{}}
\toprule

\textbf{Formation}
& \shortstack{\textbf{Time}\\\textbf{(s)}}
& \shortstack{\textbf{Height Std.}\\\textbf{(cm)}}
& \shortstack{\textbf{Tilt P95}\\\textbf{(deg)}}
& \shortstack{\textbf{Yaw RMS}\\\textbf{(deg/s)}}
& \shortstack{\textbf{Min. Clearance}\\\textbf{(cm)}} \\

\midrule

Outbound
& 122
& 1.0
& 1.30
& 7.9
& 9.6 \\

Return
& 119
& 1.3
& 2.12
& 7.3
& 9.1 \\

\bottomrule
\end{tabular*}
\end{table}

\begin{figure}[t]
    \centering
    \includegraphics[width=\linewidth]{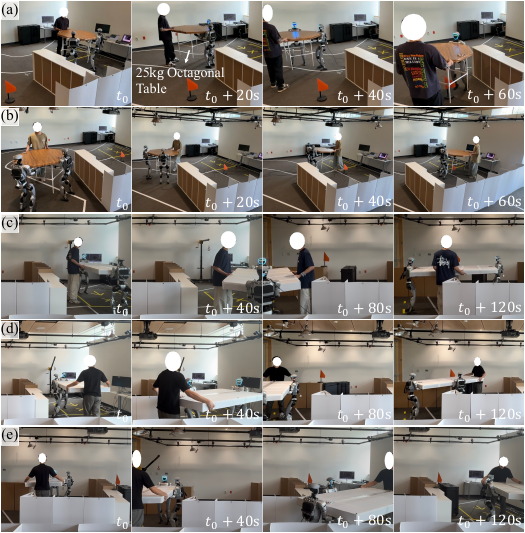}
    \caption{Representative snapshots from five real-world human--robot cooperative-transport trials showing temporal progression across five task configurations, including transport of an octagonal table.}
    \label{fig:real_world_five_cases}
\end{figure}

\subsection{Mechanism and Ablation Analysis}
\label{sec:mechanism_ablations}

Fig.~\ref{fig:sail_talo_ablation}(a)--(b) isolates SAIL under the same exact joint CBF-QP. Compared with HAPPO+CBF, SAIL+CBF reduces raw-tree violation
by $48.8\%$ and proposal--execution correction by $85.2\%$. Under the deployment setting, these improvements remain in the learned tree: the violation rate decreases from $0.29$ to $0.21$, and the exact-CBF correction $\|\mathcal Q_{\mathrm{CBF}}(x,a_{\mathrm{ST}})-a_{\mathrm{ST}}\|_2$ decreases from $0.21$ to $0.13$. This persistence indicates partial internalization of the enforced safety constraints by the nominal policy.

Fig.~\ref{fig:sail_talo_ablation}(c)--(d) evaluates the team-aware update regulation by comparing SAIL+CBF with SAGE on the same tree--SAIL proposal graph. TALO reduces the update-consistency gap by $50.8\%$ and raises cosine alignment
from $0.901$ to $0.926$. These quantities diagnose agreement with the team-reference update rather than convergence. The Lyapunov correction attenuates discrepancy-increasing updates; because TALO targets update consistency rather than proposal violation, SAGE need not improve every SAIL-specific diagnostic monotonically.

Fig.~\ref{fig:payload_panels} further compares payload tilt, heading error, height error, and completion time. Together with Table~\ref{tab:global_performance}, these results show that task success is recovered without relaxing the common safety filter, but by changing how the tree policies are trained around the fixed constrained-execution layer.

\subsection{Auditability and Physical Validation}
\label{sec:auditability}

Hardware trials use two Unitree G1 robots and a human partner. Fig.~\ref{fig:real_world_three_cases} presents Vicon trajectories and payload velocity profiles from four trials per formation.

The joint leaf tuple $\boldsymbol{\ell}_t=(\ell_{1,t},\ell_{2,t},\ell_{3,t})$ exhibits recurring route-dependent patterns across episodes in Fig.~\ref{fig:colortrace}. Table~\ref{tab:leaf_interpretability} shows higher purity and NMI for joint tuples than individual leaves, while both representations depart from the shuffled null. Purity and NMI measure association with discretized route progress, while the shuffle $z$-score compares the observed association with a label-permuted null. This stronger joint association is consistent with coordinated specialization among independently parameterized local trees. Fig.~\ref{fig:real_world_five_cases} shows octagonal and rectangular payloads transported through confined passages. Table~\ref{tab:realworld_formations} summarizes transport efficiency, payload stability, and obstacle clearance, supporting feasibility in the tested hardware configurations.

\begin{figure}[t]
    \centering
    \includegraphics[width=\columnwidth]{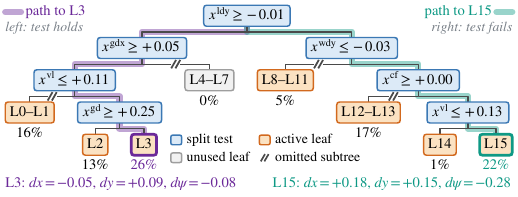}
    \caption{Representative decision traces through a deployed hard-tree policy in the Isaac Lab Maze, with green and purple branches marking two complete root-to-leaf predicate sequences and their affine leaf commands; unselected sibling branches are collapsed for readability.}
    \label{fig:tree}
\end{figure}

Fig.~\ref{fig:tree} traces two representative decisions from one deployed robot tree using observations recorded during an Isaac Lab Maze rollout. The predicates are expressed in the robot body frame, and leaf actions are ordered as $[\Delta x,\Delta y,\Delta\psi]$. The green route reaches $L_{15}$, which appears near a corner-entry regime. Its action $[0.18,0.15,-0.28]$ combines forward motion, leftward translation, and yaw rotation. The purple route reaches $L_3$ near the terminal corridor; its lower-magnitude action $[-0.05,0.09,-0.08]$ is consistent with convergent final positioning. The nominal decision layer directly exposes the active predicate sequence and affine command, while the CBF-QP may separately modify the executed action.

\section{Conclusions}

This paper presented SAGE, a training framework for safety-constrained multi-agent human--robot collaboration with auditable hard-tree nominal policies and CBF execution. By addressing the proposal--execution mismatch and the independent--team update mismatch, SAGE aligns policy learning with the constrained controller retained at deployment. The main conclusions are summarized as follows:

\begin{itemize}
    \item The safety--performance tradeoff introduced by CBF filtering is mitigated by SAGE. The exact CBF-QP reduced collisions by 98.5\% but lowered success from 67.3\% to 59.3\%; SAGE aligned nominal-policy training with constrained execution, raising success to 71.0\% while reducing collisions to 0.5 steps per thousand environment steps.

    \item SAIL and TALO address the two learning mismatches. SAIL uses a finite-penalty proposal map to preserve constraint-normal sensitivity, reducing proposal violation by 48.8\% and proposal--execution correction by 85.2\%. TALO builds a team-aware update reference from team-averaged context and joint proposal likelihood, then applies a Lyapunov half-space correction, reducing the update-consistency gap by 50.8\%.

    \item Explainable hard-tree policies expose explicit routing predicates and local actions, while the exact joint CBF-QP enforces the specified execution constraints. Joint leaf tuples achieved 0.845 route-progress purity and 0.78 NMI, and experiments with two Unitree G1 humanoids and a human partner demonstrated the architecture in physical collaboration.
\end{itemize}

\bibliographystyle{IEEEtran}
\bibliography{IEEEexample}

@article{ajoudani2018progress,
  author    = {Ajoudani, Arash and Zanchettin, Andrea Maria and
               Ivaldi, Serena and Albu-Sch{\"a}ffer, Alin and
               Kosuge, Kazuhiro and Khatib, Oussama},
  title     = {Progress and Prospects of the Human--Robot Collaboration},
  journal   = {Autonomous Robots},
  volume    = {42},
  number    = {5},
  pages     = {957--975},
  year      = {2018},
  doi       = {10.1007/s10514-017-9677-2}
}

@inproceedings{dragan2013legibility,
  author    = {Dragan, Anca D. and Lee, Kenton C. T. and
               Srinivasa, Siddhartha S.},
  title     = {Legibility and Predictability of Robot Motion},
  booktitle = {Proceedings of the 8th ACM/IEEE International
               Conference on Human-Robot Interaction},
  pages     = {301--308},
  year      = {2013},
  doi       = {10.1109/HRI.2013.6483603}
}

@misc{Cognition2Control,
  author        = {Zhang, Hao and Zhao, Ding and Tseng, H. Eric},
  title         = {Cognition to Control -- Multi-Agent Learning for
                   Human-Humanoid Collaborative Transport},
  year          = {2026},
  eprint        = {2603.03768},
  archivePrefix = {arXiv},
  primaryClass  = {cs.RO}
}

@inproceedings{bernardtiong2024cooperative,
  author    = {Bernard-Tiong, Ing-Sheng and Tsurumine, Yoshihisa and
               Sota, Ryosuke and Shibata, Kazuki and Matsubara, Takamitsu},
  title     = {Cooperative Grasping and Transportation Using Multi-Agent
               Reinforcement Learning with Ternary Force Representation},
  booktitle = {2025 IEEE/SICE International Symposium on System Integration},
  pages     = {973--978},
  year      = {2025},
  doi       = {10.1109/SII59315.2025.10871115}
}

@inproceedings{yu2022surprising,
  author    = {Yu, Chao and Velu, Akash and Vinitsky, Eugene and
               Gao, Jiaxuan and Wang, Yu and Bayen, Alexandre and Wu, Yi},
  title     = {The Surprising Effectiveness of PPO in Cooperative
               Multi-Agent Games},
  booktitle = {Advances in Neural Information Processing Systems},
  volume    = {35},
  pages     = {24611--24624},
  year      = {2022}
}

@inproceedings{kuba2022trust,
  author    = {Kuba, Jakub Grudzien and Chen, Ruiqing and Wen, Muning and
               Wen, Ying and Sun, Fanglei and Wang, Jun and Yang, Yaodong},
  title     = {Trust Region Policy Optimisation in Multi-Agent
               Reinforcement Learning},
  booktitle = {International Conference on Learning Representations},
  year      = {2022}
}

@article{zhong2024harl,
  author  = {Zhong, Yifan and Kuba, Jakub Grudzien and Feng, Xidong and
             Hu, Siyi and Ji, Jiaming and Yang, Yaodong},
  title   = {Heterogeneous-Agent Reinforcement Learning},
  journal = {Journal of Machine Learning Research},
  volume  = {25},
  number  = {32},
  pages   = {1--67},
  year    = {2024}
}

@inproceedings{bastani2018viper,
  author    = {Bastani, Osbert and Pu, Yewen and Solar-Lezama, Armando},
  title     = {Verifiable Reinforcement Learning via Policy Extraction},
  booktitle = {Advances in Neural Information Processing Systems},
  volume    = {31},
  year      = {2018}
}

@inproceedings{milani2022maviper,
  author    = {Milani, Stephanie and Zhang, Zhicheng and Topin, Nicholay and
               Shi, Zheyuan Ryan and Kamhoua, Charles A. and
               Papalexakis, Evangelos E. and Fang, Fei},
  title     = {{MAVIPER}: Learning Decision Tree Policies for
               Interpretable Multi-Agent Reinforcement Learning},
  booktitle = {Machine Learning and Knowledge Discovery in Databases},
  pages     = {251--266},
  year      = {2022},
  publisher = {Springer},
  doi       = {10.1007/978-3-031-26412-2_16}
}

@inproceedings{paleja2022icct,
  author    = {Paleja, Rohan and Niu, Yaru and Silva, Andrew and
               Ritchie, Chace and Choi, Sugju and Gombolay, Matthew},
  title     = {Learning Interpretable, High-Performing Policies for
               Autonomous Driving},
  booktitle = {Proceedings of Robotics: Science and Systems},
  year      = {2022},
  doi       = {10.15607/RSS.2022.XVIII.068}
}

@article{liu2025mixrts,
  author  = {Liu, Zichuan and Zhu, Yuanyang and Wang, Zhi and
             Gao, Yang and Chen, Chunlin},
  title   = {{MIXRTs}: Toward Interpretable Multi-Agent Reinforcement
             Learning via Mixing Recurrent Soft Decision Trees},
  journal = {IEEE Transactions on Pattern Analysis and Machine Intelligence},
  volume  = {47},
  number  = {5},
  pages   = {4090--4107},
  year    = {2025},
  doi     = {10.1109/TPAMI.2025.3540467}
}

@article{vos2024dtpo,
  author  = {Vos, Dani{\"e}l and Verwer, Sicco},
  title   = {Optimizing Interpretable Decision Tree Policies for
             Reinforcement Learning},
  journal = {arXiv preprint arXiv:2408.11632},
  year    = {2024}
}

@article{ames2017control,
  author  = {Ames, Aaron D. and Xu, Xiangru and Grizzle, Jessy W. and
             Tabuada, Paulo},
  title   = {Control Barrier Function Based Quadratic Programs for
             Safety Critical Systems},
  journal = {IEEE Transactions on Automatic Control},
  volume  = {62},
  number  = {8},
  pages   = {3861--3876},
  year    = {2017},
  doi     = {10.1109/TAC.2016.2638961}
}

@inproceedings{cheng2019end,
  author    = {Cheng, Richard and Orosz, G{\'a}bor and
               Murray, Richard M. and Burdick, Joel W.},
  title     = {End-to-End Safe Reinforcement Learning through Barrier
               Functions for Safety-Critical Continuous Control Tasks},
  booktitle = {Proceedings of the AAAI Conference on Artificial Intelligence},
  volume    = {33},
  number    = {1},
  pages     = {3387--3395},
  year      = {2019},
  doi       = {10.1609/aaai.v33i01.33013387}
}

@article{xiao2023barriernet,
  author  = {Xiao, Wei and Wang, Tsun-Hsuan and Hasani, Ramin and
             Chahine, Makram and Amini, Alexander and Li, Xiao and
             Rus, Daniela},
  title   = {{BarrierNet}: Differentiable Control Barrier Functions
             for Learning of Safe Robot Control},
  journal = {IEEE Transactions on Robotics},
  volume  = {39},
  number  = {3},
  pages   = {2289--2307},
  year    = {2023},
  doi     = {10.1109/TRO.2023.3249564}
}

@article{zhang2025gcbfplus,
  author  = {Zhang, Songyuan and So, Oswin and Garg, Kunal and Fan, Chuchu},
  title   = {{GCBF+}: A Neural Graph Control Barrier Function Framework
             for Distributed Safe Multiagent Control},
  journal = {IEEE Transactions on Robotics},
  volume  = {41},
  pages   = {1533--1552},
  year    = {2025},
  doi     = {10.1109/TRO.2025.3530348}
}

@inproceedings{balduzzi2018mechanics,
  author    = {Balduzzi, David and Racani{\`e}re, S{\'e}bastien and
               Martens, James and Foerster, Jakob and Tuyls, Karl and
               Graepel, Thore},
  title     = {The Mechanics of n-Player Differentiable Games},
  booktitle = {Proceedings of the 35th International Conference
               on Machine Learning},
  volume    = {80},
  series    = {Proceedings of Machine Learning Research},
  pages     = {354--363},
  year      = {2018},
  publisher = {PMLR}
}

@inproceedings{foerster2018lola,
  author    = {Foerster, Jakob N. and Chen, Richard Y. and
               Al-Shedivat, Maruan and Whiteson, Shimon and
               Abbeel, Pieter and Mordatch, Igor},
  title     = {Learning with Opponent-Learning Awareness},
  booktitle = {Proceedings of the 17th International Conference on
               Autonomous Agents and MultiAgent Systems},
  pages     = {122--130},
  year      = {2018}
}

@article{mittal2023orbit,
  author  = {Mittal, Mayank and Yu, Calvin and Yu, Qinxi and
             Liu, Jingzhou and Rudin, Nikita and Hoeller, David and
             Yuan, Jia Lin and Singh, Ritvik and Guo, Yunrong and
             Mazhar, Hammad and Mandlekar, Ajay and Babich, Buck and
             State, Gavriel and Hutter, Marco and Garg, Animesh},
  title   = {Orbit: A Unified Simulation Framework for Interactive
             Robot Learning Environments},
  journal = {IEEE Robotics and Automation Letters},
  volume  = {8},
  number  = {6},
  pages   = {3740--3747},
  year    = {2023},
  doi     = {10.1109/LRA.2023.3270034}
}

@article{zhang2026halo,
  title={HALO: Learning Human-Robot Collaboration via Heterogeneous-Agent Lyapunov Policy Optimization},
  author={Zhang, Hao and Niu, Yaru and Wang, Yikai and Zhao, Ding and Tseng, H Eric},
  journal={arXiv preprint arXiv:2603.03741},
  year={2026},
  publisher={International Conference on Machine Learning (ICML)}
}

@article{zhang2025multi,
  title={Multi-scale reinforcement learning of dynamic energy controller for connected electrified vehicles},
  author={Zhang, Hao and Lei, Nuo and Li, Shengbo Eben and Zhang, Junzhi and Wang, Zhi},
  journal={IEEE Transactions on Intelligent Transportation Systems},
  year={2025},
  publisher={IEEE}
}

@article{zhang2025bi,
  title={Bi-level transfer learning for lifelong-intelligent energy management of electric vehicles},
  author={Zhang, Hao and Lei, Nuo and Peng, Wang and Li, Bingbing and Lv, Shujun and Chen, Boli and Wang, Zhi},
  journal={IEEE Transactions on Intelligent Transportation Systems},
  volume={26},
  number={10},
  pages={16174--16187},
  year={2025},
  publisher={IEEE}
}

@article{zhang2026interaction,
  title={Interaction-Aware Whole-Body Control for Compliant Object Transport},
  author={Zhang, Hao and Tseng, Yves and Zhao, Ding and Tseng, H Eric},
  journal={arXiv preprint arXiv:2603.03751},
  year={2026}
}

\end{document}